\documentclass{article} 
\usepackage{colm2024_conference}
\usepackage{microtype}
\usepackage{hyperref}
\usepackage{url}
\usepackage{booktabs}
\usepackage{CJKutf8}
\usepackage{graphicx}
\usepackage{multicol}
\usepackage{multirow}
\usepackage{array}
\usepackage{amsmath}
\usepackage{booktabs}
\usepackage{enumitem}
\usepackage{wrapfig}
\usepackage{algorithm}
\usepackage{algpseudocode}
\usepackage{microtype}
\usepackage{amsmath}
\usepackage{colortbl}
\usepackage[utf8]{inputenc}
\definecolor{lightgray}{rgb}{0.9,0.9,0.9}
\usepackage{caption}
\usepackage{subcaption}
\usepackage{xcolor}
\usepackage{setspace}
\usepackage{url}
\usepackage{colortbl}
\usepackage{tabularx}
\usepackage{blindtext}
\usepackage{pgfplots}
\pgfplotsset{compat=1.18} 
\usepackage{tikz}
\usetikzlibrary{er,positioning,bayesnet}
\usepackage{makecell}
\usepackage{tipa}
\usepackage{siunitx}
\usepackage{nicefrac}
\usepackage{tocloft}
\usepackage{listings}
\usepackage{tcolorbox}
\usepackage{xltabular}
\usepackage{adjustbox}
\usepackage{xurl}
\usepackage{setspace}
\usepackage{lipsum}

\title{Hunyuan-MT Technical Report}

\author{
	\bf \large Tencent Hunyuan Team
}
\begin{document}
\begin{CJK*}{UTF8}{gbsn}
\maketitle
		
\begin{abstract}
In this report, we introduce \textbf{Hunyuan-MT-7B}, our first open-source multilingual translation model, which supports bidirectional translation across 33 major languages and places a special emphasis on translation between Mandarin and several ethnic minority languages as well as dialects. Furthermore, to serve and address diverse translation scenarios and enhance model performance at test time, we introduce \textbf{Hunyuan-MT-Chimera-7B}, a translation model inspired by the \textit{slow thinking} mode. This model integrates multiple outputs generated by the \textbf{Hunyuan-MT-7B} model under varying parameter settings, thereby achieving performance superior to that of conventional \textit{slow-thinking} models based on Chain-of-Thought (CoT). The development of our models follows a holistic training process specifically engineered for multilingual translation, which begins with general and MT-oriented pre-training to build foundational capabilities, proceeds to Supervised Fine-Tuning (SFT) for task-specific adaptation, and culminates in advanced alignment through Reinforcement Learning (RL) and weak-to-strong RL. Through comprehensive experimentation, we demonstrate that both \textbf{Hunyuan-MT-7B} and \textbf{Hunyuan-MT-Chimera-7B} significantly outperform all translation-specific models of comparable parameter size and most of the SOTA large models, particularly on the task of translation between Mandarin and minority languages as well as dialects.

{\color{purple}\textbf{In the WMT2025 shared task (General Machine Translation), our models demonstrate state-of-the-art performance, ranking first in 30 out of 31 language pairs}}. This result highlights the robustness of our models across a diverse linguistic spectrum, encompassing high-resource languages such as Chinese, English, and Japanese, as well as low-resource languages including Czech, Marathi, Estonian, and Icelandic.
			
\small {\color{blue}\textbf{Hunyuan-MT-7B}}: \url{https://huggingface.co/tencent/Hunyuan-MT-7B}

\small {\color{blue}\textbf{Hunyuan-MT-Chimera-7B}}: \url{https://huggingface.co/tencent/Hunyuan-MT-Chimera-7B}

\normalsize {\color{blue}\textbf{Code Repository}}: \url{https://github.com/Tencent-Hunyuan/Hunyuan-MT}
\end{abstract}

\begin{figure}[h]
\centering
\includegraphics[width=1\linewidth]{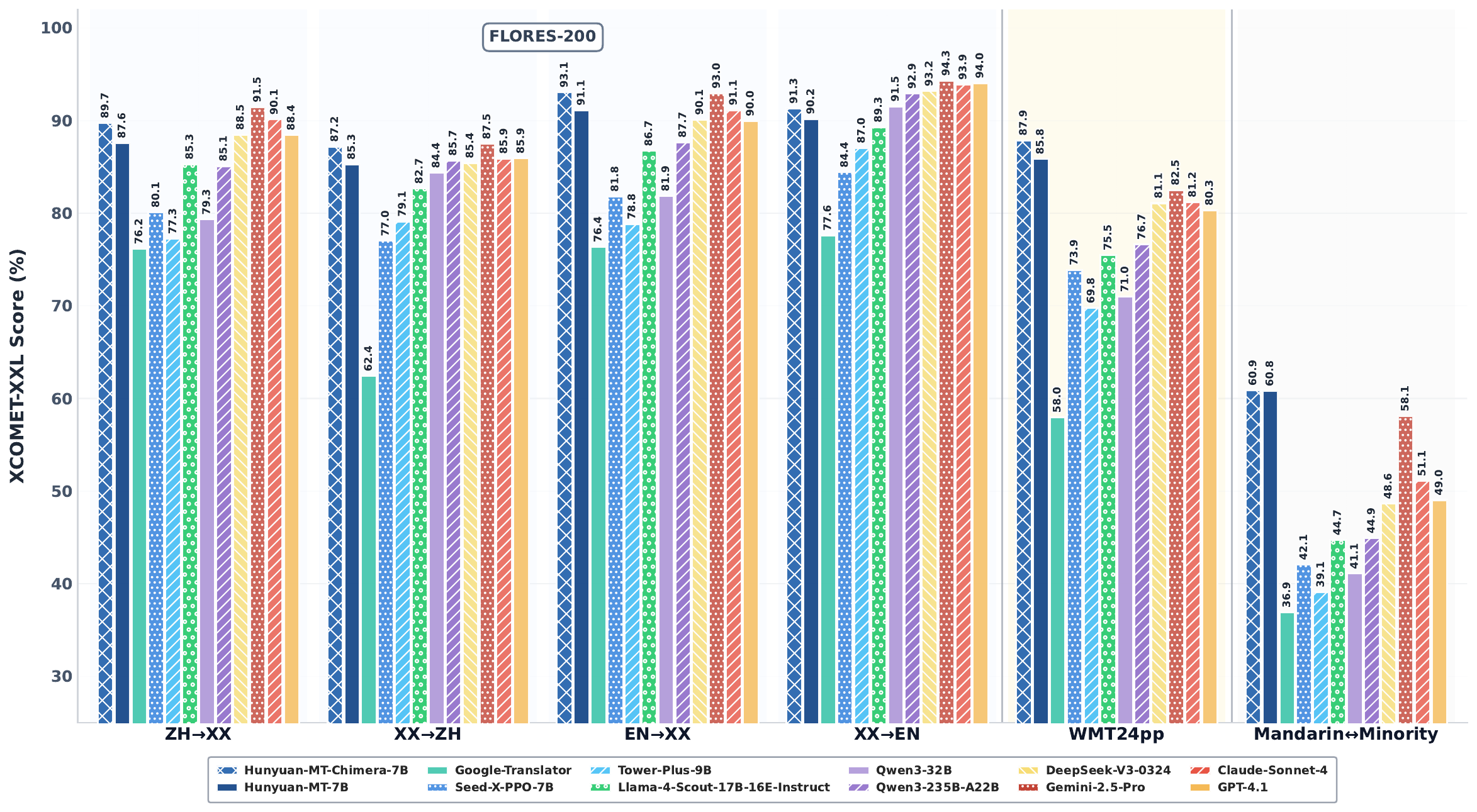}
\caption{Benchmark performance of Hunyuan-MT models and state-of-the-art baselines.}
\label{fig:overall}
\end{figure}

		\section{Introduction}
Machine Translation (MT) has emerged as both a critically important practical application and one of the most formidable research challenges that the computational linguistics community has pursued over the past several decades \citep{brown1990statistical, brown1993mathematics, bleu, sutskever2014, bahdanau2015, wu2016google, vaswani2017attention}. The recent advent and rapid advancement of Large Language Models (LLMs) have revolutionized the learning paradigm underlying MT systems, catalyzing a shift from traditional rule-based and statistical approaches toward sophisticated large-scale neural learning methodologies \citep{zhu2024,kocmi2024,pang2025}. This continuous technological evolution of LLMs has dramatically pushed the boundaries of achievable translation quality to unprecedented levels, with state-of-the-art models such as GPT-4.1 \citep{gpt41}, Gemini-2.5-Pro \citep{gemini_pro}, and Claude-Sonnet-4 \citep{claude4},  demonstrating remarkable capabilities that exceed the performance of expert human translators across specific language pairs.

Nevertheless, significant challenges persist \citep{pang2025}, particularly in translating non-literal language, such as internet neologisms, slang, and specialized terminology, as well as place names. Furthermore, a prevailing bias within MT research favors high-resource language pairs, leaving translation for low-resource and minority languages critically under-resourced. The translation between China's minority languages and Mandarin constitutes a particularly acute manifestation of this neglect. Beyond its technical dimensions, facilitating high-quality translation in this context is pivotal for promoting social inclusion, preserving cultural heritage, and ensuring equitable access to essential services and information for minority communities \citep{hu2019, Lin02102021}. Despite this pressing societal imperative, this specific domain represents a significant lacuna within the MT field.

Addressing these issues requires more than robust linguistic comprehension; it necessitates the ability to generate expressions that are both culturally resonant and idiomatically natural, thereby transcending literal word-for-word translation. Meanwhile, a notable performance disparity remains between proprietary and open-source models, a gap often attributed to the comparatively limited scale of open-source systems. This problem is compounded by a scarcity of well-defined methodologies for developing advanced LLM-based MT systems, which impedes the broader community's efforts to deploy and refine effective solutions \citep{jiao2023, pang2025, kocmi2024, seedx}.

Through extensive evaluations on representative MT benchmarks, Hunyuan-MT demonstrates superior performance, outperforming not only translation-specialized models of comparable size and prominent closed-source systems, such as Google-Translator, but also a range of larger LLMs, as detailed in Figure~\ref{fig:overall}. Furthermore, it is noteworthy that our model demonstrates significant superiority over all state-of-the-art LLMs on the task of translation between China's ethnic minority languages and Mandarin Chinese (Minority$\Leftrightarrow$Mandarin Translation).
In this technical report, we introduce Hunyuan-MT, the culmination of our ongoing efforts to develop more effective LLM-based multilingual translation models. Below, we show the main contributions of this technical report:
\begin{enumerate}[left=2pt, itemsep=2pt, topsep=4pt]
	\item \textbf{Hunyuan-MT-7B}. We present Hunyuan-MT-7B, a novel, open-source multilingual translation model with 7B parameters, which facilitates bidirectional translation among a diverse set of 33 languages. Through extensive benchmarking, our model demonstrates SOTA performance against existing models in the approximate parameter size. Furthermore, a key feature of Hunyuan-MT-7B is its robust support for low-resource language pairs, specifically enabling translation between Mandarin and several ethnic minority languages as well as dialects.
	\item \textbf{Hunyuan-MT-Chimera-7B}. We propose a new paradigm for high-quality machine translation, embodied in our model, Hunyuan-MT-Chimera-7B. As the first open-source weak-to-strong fusion model of its kind, it is architected for "slow thinking" translation tasks where quality is paramount. Unlike traditional methods that rely on a single decoding path or unstructured reasoning like CoT, our approach involves a two-stage process. First, a base model (Hunyuan-MT-7B) generates a portfolio of diverse translation candidates. Second, Hunyuan-MT-Chimera-7B, an expert model trained specifically for this purpose, synthesizes these "weaker" candidates into a single "strong" output. This learned synthesis mechanism yields translations of a quality unattainable by any individual candidate and demonstrates superior performance over contemporary CoT-based models.
	\item \textbf{A Training Recipe}. We introduce a simple yet effective training framework that progressively refines the model through a sequence of several distinct stages. Our methodology begins with foundational knowledge acquisition via general and MT-oriented pre-training, followed by task-specific adaptation using Supervised Fine-Tuning (SFT). Crucially, we then employ two successive optimization phases: a Reinforcement Learning (RL) stage and an advanced weak-to-strong RL stage.
	\item \textbf{Mandarin$\Leftrightarrow$Minority Translation}. To the best of our knowledge, this work represents the first systematic effort to optimize bidirectional translation performance of Mandarin-Kazakh, Mandarin-Uyghur, Mandarin-Mongolian, and Mandarin-Tibetan. Our empirical evaluation demonstrates that the resulting models significantly outperform leading contemporary LLMs on these translation directions, achieving the state-of-the-art performance.
\end{enumerate}
The remainder of this paper is organized as follows: we first detail the training methodology for Hunyuan-MT series models, then present experimental results for their pre-trained and post-trained variants, and finally, conclude with a discussion of key findings and future research directions.

\section{Pre-training Stage}
In this section, we describe the details of our pre-training approach and present experimental results from evaluating the base models on standard benchmarks.
\subsection{General Pre-training}
For the foundational pre-training stage, we employ a multilingual corpus that jointly trains on data from Chinese and English, supplemented by a substantial collection of low-resource languages. The multilingual data component comprises 1.3 trillion tokens, encompassing 112 languages and dialects (excluding Chinese and English) sourced from a wide array of domains.

To govern the quality of this diverse dataset, we developed and implemented a proprietary quality assessment model. This model evaluates text along three key dimensions: Knowledge Value, which measures informational density and accuracy; Authenticity, which verifies the genuineness of the content; and Writing Style, which appraises linguistic quality and coherence. Each dimension is scored on a three-point scale (0, 1, 2). A weighted composite score is then calculated, where the weighting of these dimensions is strategically adjusted based on the provenance of the data. For instance, for data sourced from academic literature, books, and professional websites, we assign a higher weight to the Knowledge Value dimension, prioritizing content that achieves the maximum score of 2.

To maintain and control for content diversity within the training data, we established a tripartite taxonomic framework for data categorization and balancing. This framework facilitates both data filtering and proportional adjustments within the training mixture:
\begin{itemize}
	\item Disciplinary Tagging System
	\begin{itemize}
		\item This system categorizes data by academic discipline, enabling a balanced distribution of subjects within the corpus.
	\end{itemize}
	\item Industry Tagging System (24 categories)
	\begin{itemize}
		\item Comprising 24 distinct categories, this system ensures comprehensive coverage across various industrial sectors.
	\end{itemize}
	
	\item Content Theme Tagging System (24 categories)
	\begin{itemize}
		\item This system serves a dual purpose: it supports broad thematic diversity while also enabling the targeted exclusion of undesirable content, such as materials related to gambling or advertising.
	\end{itemize}
\end{itemize}
The integration of this quality assessment model with the tripartite taxonomic framework constitutes a comprehensive data governance strategy. This strategy ensures the final training corpus is characterized by both high quality and broad diversity. The application of this curated data processing pipeline resulted in the pre-training of the Hunyuan-7b-Base\footnote{\url{https://huggingface.co/tencent/Hunyuan-7B-Pretrain}} model.

\subsection{MT-oriented Pre-training}
During the MT-oriented pre-training stage, we incorporate a curated mixture of monolingual and bilingual corpora. The monolingual data primarily comes from the mC4 \citep{mc4} and OSCAR \citep{oscar1,oscar2} datasets. We subject this data to a rigorous cleaning pipeline that includes language identification with fastText \footnote{\url{https://github.com/facebookresearch/fastText}}, document-level deduplication via minLSH, and quality filtering with a KenLM-based model \footnote{\url{https://github.com/kpu/kenlm}} to remove high-perplexity documents. For the bilingual data, we utilize publicly available parallel corpora, such as OPUS \citep{tiedemann2012parallel} and ParaCrawl \citep{buck-koehn:2016:WMT1}, which we filter using reference-free quality estimation metrics, including CometKiwi \citep{cometkiwi}, to ensure the selection of high-quality translation pairs.

To determine the optimal data mixture ratio, we adopt a strategy inspired by RegMix \citep{liu2025regmixdatamixtureregression}. We first conduct experiments on a smaller-scale model to fit a function that maps sampling ratios to training loss. By simulating this function, we identify the mixture that minimizes the predicted loss, and we then use this ratio for the MT-oriented pre-training stage of our final translation model.

To mitigate catastrophic forgetting, we integrate a 20\% replay of the original pre-training corpus. We also design the learning rate schedule to warm up to the peak learning rate of the initial pre-training phase and then decay to its minimum value.

\section{Post-training Stage}
Following pre-training, we aim to equip the base model with robust multilingual machine translation capabilities through Supervised Fine-Tuning (SFT), Reinforcement Learning (RL), and Weak-to-Strong RL. The central challenge lies in optimizing performance for high-resource languages using model-generated and human-annotated data while ensuring effective generalization to low-resource languages.

\begin{table*}[h]
	\small
	\centering
	\begin{tabular}{l}
		\toprule
		\textbf{Prompt Template for ZH$\Leftrightarrow$XX Translation.} \\ \midrule
		把下面的文本翻译成\texttt{<target\_language>}，不要额外解释。\\ 
		\\
		\texttt{<source\_text>}
		\\ \midrule
		\textbf{Prompt Template for XX$\Leftrightarrow$XX Translation, excluding ZH$\Leftrightarrow$XX.} \\ \midrule
		Translate the following segment into \texttt{<target\_language>}, without additional explanation. \\
		\\
		\texttt{<source\_text>}
		\\ \bottomrule
	\end{tabular}
	\caption{Examples of prompt template.}\label{tab:prompt_template_mt}
\end{table*}

\begin{figure}[t!]
	\centering
	\includegraphics[width=1\linewidth]{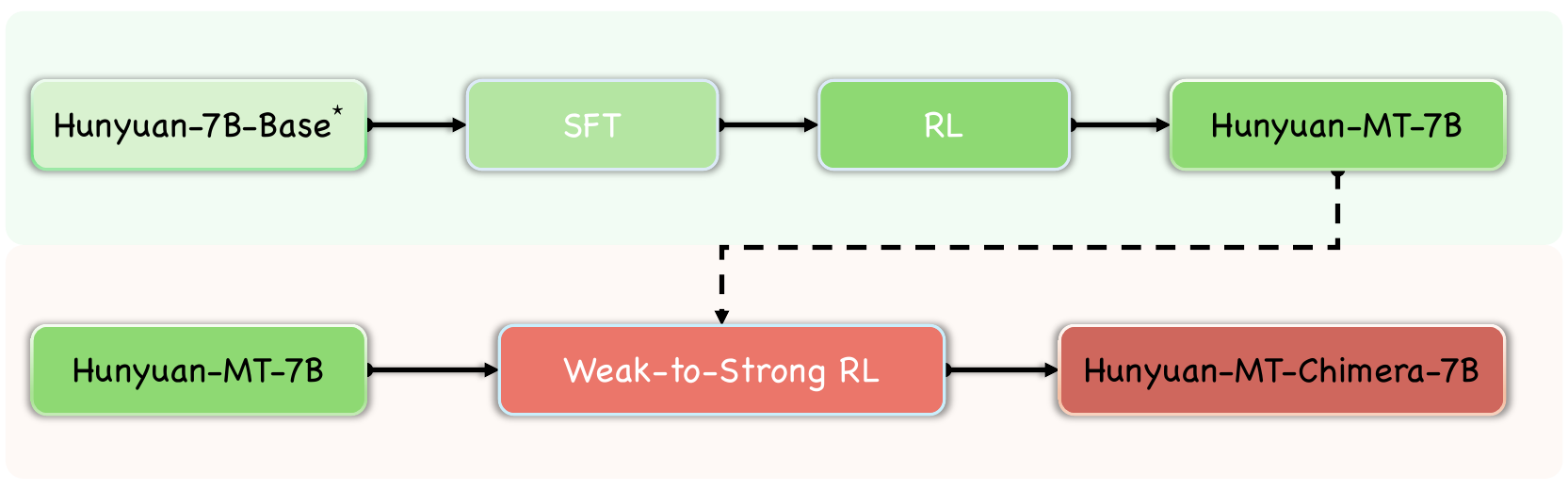}
	\caption{Post-training pipeline of the Hunyuan-MT-7B and Hunyuan-MT-Chimera-7B models.}
	\label{fig:pipline}
\end{figure}

\begin{table*}[ht]
	\centering
	\small
	\caption{Supported languages.}
	\begin{tabular}{cccccc}
		\toprule
		\textbf{Languages}  & \textbf{Abbr.} & \textbf{Languages}        & \textbf{Abbr.}  & \textbf{Languages}        & \textbf{Abbr.}\\ \midrule
		Chinese             & zh              & Malay                     & ms     & Marathi  & mr\\ 
		English               & en                    & Indonesian          & id   & Hebrew & he                 \\
		French              & fr                    & Filipino                     & tl  & Bengali & bn                  \\
		Portuguese              & pt                    & Hindi                 & hi   & Tamil & ta                 \\
		Spanish             & es                    & Traditional Chinese                    & zh-Hant   & Ukrainian & uk                 \\
		Japanese             & ja                    & Polish                & pl   & Tibetan & bo                 \\
		Turkish             & tr                    & Czech                  & cs    & Kazakh & kk              \\
		Russian              & ru                    & Dutch                   & nl   & Mongolian & mn                             \\
		Arabic            & ar                    & Khmer                   & km    & Uyghur & ug               \\
		Korean         & ko                    &  Burmese                      & my   & Cantonese & yue  \\
		Thai          & th                    & Persian                   & fa         & &           \\
		Italian             & it                    & Gujarati                 & gu    & &      \\
		German            & de                    & Urdu                & ur            & &        \\
		Vietnamese              & vi                    & Telugu                   & te    & &                \\
		\bottomrule
	\end{tabular}
	\label{tab:support_languages}
\end{table*}

\subsection{Supervised Fine-Tuning}
We adopt a two-stage SFT framework designed to progressively enhance the model's translation capabilities and its ability to follow instructions. The initial stage is dedicated to establishing a robust foundation by training the model on a comprehensive parallel corpus of approximately 3 million pairs. This dataset is aggregated from diverse sources, including established benchmarks like the Flores-200 development set and past WMT test sets, a human-annotated collection of Mandarin-centric minority language pairs, and a synthetically generated corpus from DeepSeek-V3-0324 \citep{dsv3}. To improve instruction-following, we also incorporate a 20\% component of general-purpose and MT-oriented instruction data (see an example in Table~\ref{tab:prompt_template_mt}). To maintain a high standard of data quality, we apply a filtering process using the reference-free metrics CometKiwi and GEMBA \citep{gembda}, discarding samples that fall below a specified quality threshold. Notably, the GEMBA scoring leverages the DeepSeek-V3-0324 model as the evaluator.

Building upon this foundation, the second stage aims to refine the model's translation performance further using a smaller, higher-fidelity dataset of approximately 268,000 pairs. The data for this stage undergoes a more stringent selection process. Drawing inspiration from recent work \citep{many-shot-icl, many-shot-icl-judge}, we employ many-shot in-context learning as a mechanism to vet further and select high-quality training instances. To maximize data reliability, any samples that demonstrate significant score inconsistencies across multiple evaluation rounds are flagged for manual annotation and verification by human experts. This two-stage approach allows us first to build broad multilingual capabilities and then hone them with a meticulously curated dataset.

\subsection{Reinforcement Learning}
While large-scale RL has demonstrated significant effectiveness in enhancing reasoning capabilities for tasks with structured outputs, such as mathematical problem-solving and code generation \citep{ds-r1, deepcoder2025, fastcurl, kimik2}, its application to MT presents a unique challenge. Unlike in structured domains, MT outputs are characterized by considerable semantic diversity, which makes them resistant to evaluation through explicit, rule-based evaluation. 

To address this challenge, we adopt GRPO \citep{grpo} as the RL algorithm and design a comprehensive reward function comprising the following components:

\begin{itemize}
	\item \noindent\textbf{Quality-Aware Reward}. To ensure translation quality during RL training, we employ two complementary reward signals. The first is XCOMET-XXL, a widely adopted metric in translation evaluation scenarios that demonstrates high correlation with human assessments. The second reward utilizes DeepSeek-V3-0324 for scoring, with prompts adapted from the GEMBA framework.
	\item \noindent\textbf{Terminology-Aware Reward}. While XCOMET-based rewards primarily focus on overall semantic similarity between translated outputs and reference translations, they may inadequately capture critical information such as domain-specific terminology. To address this limitation, we incorporate the word alignment-based reward metric proposed in TAT-R1 \citep{TAT-R1}. This reward mechanism extracts key information, including terminology, through word alignment tools, then computes the overlap ratio of these critical elements between the translation output and reference. Higher overlap ratios yield greater rewards, thereby enhancing the model's attention to terminology and other crucial information during training.
	\item \noindent\textbf{Repetition Penalty}. We observe that models tend to generate repetitive outputs in later stages of reinforcement training, potentially leading to training collapse. To mitigate this issue, we implement a repetition detection mechanism that applies penalties when repetitive patterns are identified, thereby maintaining output diversity and training stability.
\end{itemize}

\subsection{Weak-to-Strong RL}
Recent studies demonstrate that increasing inference time significantly enhances model performance on mathematical and coding tasks \citep{test-time-scaling, test-time-scaling-survey}. However, initial experiments incorporating Chain-of-Thought (CoT) \citep{cot} reasoning into translation tasks yield limited improvements in translation quality, as discussed comprehensively in Section 6. Consequently, in this report, we explore a new test-time scaling methodology from a novel perspective to enhance test-time performance for MT.

Specifically, to address the above issue, we propose a weak-to-strong RL approach that generates multiple translation outputs and employs a fusion model based on Hunyuan-MT-7B to aggregate these outputs through GRPO. The reward function comprises three primary elements: XCOMET-XXL scoring, DeepSeek-V3-0324 scoring, and a repetition penalty term. This multi-faceted reward mechanism ensures comprehensive evaluation of translation quality while mitigating redundancy in generated outputs. This methodology culminates in the development of our Hunyuan-MT-7B-Chimera model. Specifically, the prompt template is shown in Table~\ref{tab:prompt_template_chimera}.

During test-time inference, Hunyuan-MT-7B-Chimera accepts multiple translation candidates as input and synthesizes their respective strengths to produce a superior unified translation output. This aggregation approach leverages the complementary advantages of diverse translation hypotheses to achieve enhanced translation quality.

\begin{table*}[t!]
	\small
	\centering
	\begin{tabular}{l}
		\toprule
		\textbf{Prompt Template for Hunyuan-MT-Chimera-7B. } \\ \midrule
		Analyze the following multiple \texttt{<target\_language>} translations of the \texttt{<source\_language>} \\
		segment surrounded in triple backticks and generate a single refined \texttt{<target\_language>} \\
		translation. Only output the refined translation, do not explain. \\
		\\
		The \texttt{<source\_language>} segment: \\
		\texttt{\textasciigrave\textasciigrave\textasciigrave<source\_text>\textasciigrave\textasciigrave\textasciigrave} \\
		\\
		The multiple \texttt{<target\_language>} translations: \\
		1. \texttt{\textasciigrave\textasciigrave\textasciigrave<translated\_text1>\textasciigrave\textasciigrave\textasciigrave} \\
		2. \texttt{\textasciigrave\textasciigrave\textasciigrave<translated\_text2>\textasciigrave\textasciigrave\textasciigrave} \\
		3. \texttt{\textasciigrave\textasciigrave\textasciigrave<translated\_text3>\textasciigrave\textasciigrave\textasciigrave} \\
		4. \texttt{\textasciigrave\textasciigrave\textasciigrave<translated\_text4>\textasciigrave\textasciigrave\textasciigrave} \\
		5. \texttt{\textasciigrave\textasciigrave\textasciigrave<translated\_text5>\textasciigrave\textasciigrave\textasciigrave} \\
		6. \texttt{\textasciigrave\textasciigrave\textasciigrave<translated\_text6>\textasciigrave\textasciigrave\textasciigrave} \\
		\\ \midrule
		
	\end{tabular}
	\caption{Prompt template of Hunyuan-MT-Chimera-7B.}\label{tab:prompt_template_chimera}
\end{table*}

\section{Experiments}

\subsection{Benchmarks}
We conduct comprehensive evaluations of the base model of the Hunyuan-MT series, focusing primarily on its performance in general knowledge, reasoning, mathematics, scientific knowledge, coding, and multilingual capabilities. Specifically, the evaluation benchmarks for pre-trained models include nine widely used benchmarks: MMLU-Pro \citep{MMLU-Pro}, SuperGPQA \citep{SuperGPQA}, BBH \citep{BBH}, GPQA \citep{GPQA}, GSM8K \citep{GSM8K}, MATH \citep{MATH}, MultiPL-E \citep{MultiPL-E}, CRUX-O of CRUXEval \citep{CRUX-O}, and INCLUDE \citep{INCLUDE}.

To comprehensively evaluate the multilingual translation capabilities, we conducted extensive experiments using the following test sets:
\begin{itemize}
	\item \textbf{Flores-200}\footnote{\url{https://huggingface.co/datasets/Muennighoff/flores200}} \citep{Flores-200}. We select 1,056 language pairs across 33 different languages (detailed in the Appendix) from the Flores-200 dataset. These pairs are systematically categorized into five groups: English$\Rightarrow$XX, XX$\Rightarrow$English, Chinese$\Rightarrow$XX, XX$\Rightarrow$Chinese, and XX$\Rightarrow$XX translations.
	\item \textbf{WMT24pp}\footnote{\url{https://huggingface.co/datasets/google/wmt24pp}} \citep{wmt24pp}. We incorporate development sets from WMT-25, encompassing English-to-XX translations across 25 target languages. WMT24pp serves as the official development set recommended by WMT25. We select 29 language pairs that overlap with the general translation track of the WMT25, with a primary focus on English-to-XX directions.
	\item \textbf{Mandarin$\Leftrightarrow$Minority Testset}. This test set encompasses translations between Chinese and minority languages: Tibetan, Mongolian, Uyghur, and Kazakh.
\end{itemize}

\subsection{Evaluation Metrics}
We evaluate translations using two complementary approaches: automatic metrics and human evaluation. For automatic evaluation, we use the neural metrics XCOMET-XXL \citep{xcomet} and CometKiwi \citep{cometkiwi}, which generally correlate with human judgments but can be unreliable for certain translation phenomena. To address these limitations, we conduct human evaluation in which multilingual experts rate translations on a 0–4 scale, focusing on pre-annotated error-prone points and considering accuracy, fluency, and idiomaticity.

\begin{table*}[h!]
	\centering
	\caption{Performances of state-of-the-art models on Flores-200, WMT-24pp, and Mandarin$\Leftrightarrow$Minority translation. Specifically, we report the Chinese-centric (ZH$\Rightarrow$XX and XX$\Rightarrow$ZH), English-centric (EN$\Rightarrow$XX and XX$\Rightarrow$EN), XX$\Rightarrow$XX, and Mand.$\Leftrightarrow$Min. performances of Hunyuan-MT-7B, Hunyuan-MT-Chimera-7B, and prominent existing systems. Here, Mand.$\Leftrightarrow$Min. denotes Mandarin$\Leftrightarrow$Minority translation. Models with open-source weights are marked with $^\dagger$. Baselines are categorized into three groups: (1) {\color{blue}ultra-large general models}, (2) {\color{orange}medium to small-sized general models}, and (3) {\color{purple}translation-specialized models}.}
	\scriptsize
	\renewcommand{\arraystretch}{1}
	\setlength{\tabcolsep}{5pt}
	\begin{tabular}{llccccccc}
		\toprule
		\multirow{2}{*}{\textbf{Models}} & \multirow{2}{*}{\textbf{Metrics}} & \multicolumn{5}{c}{\textbf{Flores-200}} & \multirow{2}{*}{\textbf{WMT24pp}} & \multirow{2}{*}{\textbf{Mand.$\Leftrightarrow$Min.}} \\
		\cmidrule(lr){3-7}  
		& & \textbf{ZH $\Rightarrow$ XX} & \textbf{XX $\Rightarrow$ ZH} & \textbf{EN $\Rightarrow$ XX} & \textbf{XX $\Rightarrow$ EN} & \textbf{XX $\Rightarrow$ XX} & &\\
		\midrule
		{\color{blue}GPT4.1}
		& XCOMET-XXL & 0.8843 & 0.8593 & 0.8996 & 0.9405 & 0.8258 & 0.8032 & 0.4904 \\
		\cite{gpt41} & CometKiwi  & 0.7859 & 0.7725 & 0.8702 & 0.8730 & 0.7424 & 0.7688 & 0.5020  \\
		\midrule
		{\color{blue}Claude-Sonnet-4}
		& XCOMET-XXL & 0.9013 & 0.8590 & 0.9114 & 0.9390 & 0.8548 & 0.8120 & 0.5111 \\
		\cite{claude4} & CometKiwi  & 0.7883 & 0.7739 & 0.8742 & 0.8732 & 0.7668 & 0.7804 & 0.5033 \\
		\midrule
		{\color{blue}Gemini-2.5-Pro}
		& XCOMET-XXL & 0.9146 & 0.8748 & 0.9295 & 0.9432 & 0.8773 & 0.8250 & 0.5811 \\
		\cite{gemini_pro} & CometKiwi  & 0.7859 & 0.7828 & 0.8869 & 0.8720 & 0.7674 & 0.7876 & 0.5418 \\
		\midrule
		\color{blue}{DeepSeek-V3-0324$^\dagger$}
		& XCOMET-XXL & 0.8848 & 0.8542 & 0.9010 & 0.9319 & 0.8082 & 0.8109 & 0.4865 \\
		\cite{dsv3} & CometKiwi  & 0.7722 & 0.7708 & 0.8684 & 0.8703 & 0.7450 & 0.7840 & 0.4839 \\
		\midrule
		\multirow{2}{*}{{\color{purple}Google-Translator}}
		& XCOMET-XXL & 0.7615 & 0.6243 & 0.7638 & 0.7761 & 0.6225 & 0.5796 & 0.3692 \\
		& CometKiwi  & 0.6647 & 0.5691 & 0.7242 & 0.7863 & 0.5947 & 0.5454 & 0.2891 \\
		\midrule
		{\color{purple}Tower-Plus-9B$^\dagger$}
		& XCOMET-XXL & 0.7726 & 0.7912 & 0.7884 & 0.8704 & 0.6608 & 0.6977 & 0.3912 \\
		\cite{towerplus}  & CometKiwi  & 0.6633 & 0.7429 & 0.7576 & 0.8475 & 0.6141 & 0.6741 & 0.3466 \\
		\midrule
		{\color{purple}Tower-Plus-72B$^\dagger$}
		& XCOMET-XXL & 0.7703 & 0.8235 & 0.7829 & 0.9002 & 0.7002 & 0.7276 & 0.3855 \\
		\cite{towerplus} & CometKiwi  & 0.6795 & 0.7569 & 0.7603 & 0.8624 & 0.6553 & 0.7071 & 0.3540 \\
		\midrule
		{\color{purple}Seed-X-PPO-7B$^\dagger$}
		& XCOMET-XXL & 0.8010 & 0.7702 & 0.8181 & 0.8442 & 0.6896 & 0.7388 & 0.4206 \\
		\cite{seedx} & CometKiwi  & 0.7089 & 0.7201 & 0.8118 & 0.8202 & 0.6436 & 0.7297 & 0.4861 \\
		\midrule
		{\color{purple}GemmaX2-28-9B-v0.1$^\dagger$}& XCOMET-XXL & 0.8687 & 0.8280 & 0.8806 & 0.9108 & 0.8119 & 0.7173 & 0.4269 \\ \cite{gemmaX2-28-9B} & CometKiwi  & 0.7604 & 0.7595 & 0.8576 & 0.8635 & 0.7371 & 0.7242 & 0.5178 \\
		\midrule
		{\color{purple}Gemma-3-12B-IT$^\dagger$}
		& XCOMET-XXL & 0.8567 & 0.8249 & 0.8781 & 0.9189 & 0.8020 & 0.7527 & 0.4280 \\
		\cite{gemma3} & CometKiwi  & 0.7603 & 0.7582 & 0.8498 & 0.8666 & 0.7320 & 0.7329 & 0.4186 \\
		\midrule
		{\color{purple}Gemma-3-27B-IT$^\dagger$}
		& XCOMET-XXL & 0.8783 & 0.8441 & 0.9036 & 0.9331 & 0.8381 & 0.7742 & 0.4558 \\
		\cite{gemma3} & CometKiwi  & 0.7667 & 0.7718 & 0.8630 & 0.8646 & 0.7471 & 0.7577 & 0.4844 \\
		\midrule
		{\color{orange}Qwen3-8B$^\dagger$}
		& XCOMET-XXL & 0.7250 & 0.8056 & 0.7468 & 0.8825 & 0.6544 & 0.6532 & 0.3737 \\
		\cite{qwen3} & CometKiwi  & 0.6605 & 0.7507 & 0.7257 & 0.8521 & 0.6285 & 0.6344 & 0.3166 \\
		\midrule
		{\color{orange}Qwen3-14B$^\dagger$}
		& XCOMET-XXL & 0.7826 & 0.8318 & 0.8027 & 0.9049 & 0.7228 & 0.6983 & 0.3944 \\
		\cite{qwen3} & CometKiwi  & 0.7116 & 0.7674 & 0.7877 & 0.8639 & 0.6887 & 0.6839 & 0.3568 \\
		\midrule
		{\color{orange}Qwen3-32B$^\dagger$}
		& XCOMET-XXL & 0.7933 & 0.8436 & 0.8186 & 0.9154 & 0.7433 & 0.7099 & 0.4110 \\
		\citep{qwen3} & CometKiwi  & 0.7139 & 0.7719 & 0.8001 & 0.8657 & 0.6965 & 0.6930 & 0.3841 \\
		\midrule
		{\color{orange}Qwen3-235B-A22B$^\dagger$}
		& XCOMET-XXL & 0.8509 & 0.8569 & 0.8765 & 0.9292 & 0.8018 & 0.7665 & 0.4493 \\
		\cite{qwen3} & CometKiwi  & 0.7551 & 0.7750 & 0.8475 & 0.8696 & 0.7313 & 0.7465 & 0.4456 \\
		\midrule
		{\color{orange}Llama-3.1-8B-Instruct$^\dagger$}& XCOMET-XXL & 0.6385 & 0.5148 & 0.6848 & 0.6412 & 0.4408 & 0.5130 & 0.3016 \\
		\cite{llama-3.1-8B-Instruct} & CometKiwi  & 0.6581 & 0.6185 & 0.7234 & 0.7746 & 0.5782 & 0.5990 & 0.2944 \\
		\midrule
		{\color{orange}Llama-4-Scout-17B-16E-$^\dagger$} & XCOMET-XXL & 0.8529 & 0.8266 & 0.8673 & 0.8929 & 0.7788 & 0.7550 & 0.4472 \\
		Instruct \cite{llama4_scout} & CometKiwi  & 0.7457 & 0.7417 & 0.8418 & 0.8523 & 0.7275 & 0.7422 & 0.4646 \\
		\midrule
		\multirow{2}{*}{\textbf{Hunyuan-MT-7B$^\dagger$}}
		& XCOMET-XXL & 0.8758 & 0.8528 & 0.9112 & 0.9018 & 0.7829 & 0.8585 & 0.6082 \\
		& CometKiwi  & 0.7963 & 0.7863 & 0.8742 & 0.8477 & 0.7210 & 0.8061 & 0.4162 \\
		\midrule
		\multirow{2}{*}{\textbf{Hunyuan-MT-Chimera-7B$^\dagger$}}
		& XCOMET-XXL & 0.8974 & 0.8719 & 0.9306 & 0.9132 & 0.8268 & 0.8787 & 0.6089 \\
		& CometKiwi  & 0.8066 & 0.7914 & 0.8849 & 0.8514 & 0.7424 & 0.8129 & 0.4417 \\
		\bottomrule
		\label{fig: performance}
	\end{tabular}
\end{table*}

\subsection{Main Results}
As presented in Table~\ref{fig: performance}, our experimental results indicate that the proposed Hunyuan-MT-7B and Hunyuan-MT-Chimera-7B models achieve competitive performance across the XCOMET-XXL and CometKiwi evaluation metrics. On the WMT24pp benchmark, Hunyuan-MT-7B obtains an XCOMET-XXL score of 0.8585. This result outperforms strong baselines, including large-scale models such as Gemini-2.5-Pro (0.8250) and Claude-Sonnet-4 (0.8120).

A noteworthy observation is the models' performance on Mandarin⇔Minority language translation pairs. In this setting, both Hunyuan-MT-7B (0.6082) and Hunyuan-MT-Chimera-7B (0.6089) yield scores that are considerably higher than the evaluated baselines. For instance, the scores represent a relative improvement of approximately 4.7\% over the next-best performing system, Gemini-2.5-Pro (0.5811), and show substantial gains when compared to several models specialized for translation.

When compared with dedicated translation models, our Hunyuan-MT series demonstrates promising results. The models achieve higher scores than Google-Translator and the Tower-Plus series across multiple evaluation settings, while being considerably more parameter-efficient than the 72B variant of Tower-Plus. The performance also remains competitive against other recently developed translation-optimized models, such as the Gemma-3 series.

Furthermore, the Hunyuan-MT-Chimera-7B variant exhibits significant improvements over its base model, achieving an average gain of 2.3\% in XCOMET-XXL scores across all directions in the Flores-200 dataset. The most pronounced improvements are observed in the ZH$\Rightarrow$XX (a 2.5\% increase) and XX$\Rightarrow$XX (a 5.6\% increase) translation directions. These findings suggest that our proposed weak-to-strong method is effective in further refining translation quality, enhancing performance on high-resource languages while maintaining strong capabilities on diverse slow-thinking scenarios.

\subsection{Ablation Study}
To validate the efficacy of our proposed two-stage pre-training methodology, we conduct a series of ablation studies. We first evaluate the performance of our base model, Hunyuan-7B-Base, resulting from the general pre-training phase. Subsequently, we assess the impact of the MT-oriented pre-training stage by comparing the specialized model Hunyuan-7B-Base$^\star$ against a strong baseline.

The results from our general pre-training stage, presented in Table~\ref{tab:pretrain_stage1}, confirm the efficacy of our data curation strategy. Hunyuan-7B-Base establishes new state-of-the-art performance on knowledge-intensive benchmarks such as MMLU-Pro and BBH. This outcome highlights the success of our quality assessment model in creating a corpus rich in factual and reasoning content. The model's capabilities are particularly pronounced in mathematical reasoning, where it achieves a score of 74.85 on the MATH benchmark, creating a substantial 14-point lead over the next-best competitor. 
\begin{table}[h]
	\small
	\centering
	\renewcommand{\arraystretch}{1.2}
	\setlength{\tabcolsep}{11pt}
	\caption{\textbf{Comparison among Hunyuan-7B-Base and other strong open-source baselines. The highest and second-best scores are shown in \textbf{bold} and \underline{underlined}, respectively.}}
	\label{tab:pretrain_stage1}
	\begin{tabular}{@{}lcccc@{}}
		\toprule
		& \textbf{Llama-3-8B-Base} & \textbf{Qwen2.5-7B-Base} & \textbf{Qwen3-8B-Base} & \textbf{Hunyuan-7B-Base} \\
		\midrule
		MMLU-Pro & 35.36 & 45.00 & \underline{56.73} & \textbf{57.79} \\
		SuperGPQA & 20.54 & 26.34 & \textbf{31.64} & \underline{30.47} \\
		BBH & 57.70 & 70.40 & \underline{78.40} & \textbf{82.95} \\
		\midrule
		\midrule
		GPQA & 25.80 & 36.36 & \textbf{44.44} & \underline{44.07} \\
		GSM8K & 55.30 & 85.36 & \textbf{89.84} & \underline{88.25} \\
		MATH & 20.50 & 49.80 & \underline{60.80} & \textbf{74.85} \\
		\midrule
		\midrule
		MultiPL-E & 31.45 & 50.73 & \underline{58.75} & \textbf{60.41} \\
		CRUX-O & 36.80 & 48.50 & \textbf{62.00} & \underline{60.75} \\
		\midrule
		\midrule
		IINCLUDE & 44.94 & 53.98 & \underline{59.40} & \textbf{59.55} \\
		\bottomrule
	\end{tabular}
\end{table}

The subsequent MT-oriented pre-training phase yields dramatic improvements in translation capabilities, as shown in Table~\ref{tab:pretrain_stage2}. Our specialized model, Hunyuan-7B-Base$^\star$, establishes a commanding lead over the strong Qwen3-8B-Base baseline across all machine translation benchmarks. This superiority is particularly evident on the challenging Flores-200 and Mandarin$\Leftrightarrow$Minority language benchmarks, where our model's significant gains directly validate the impact of incorporating 1.3 trillion tokens of low-resource language data. The model's advantage extends to high-resource scenarios as well, with a substantial performance increase on the WMT24pp benchmark. 

\begin{table}[h]
	\small
	\centering
	\renewcommand{\arraystretch}{1.2}
	\setlength{\tabcolsep}{3pt}
	\caption{\textbf{Comparison among Hunyuan-7B-Base$^\star$ and Qwen3-8B-Base. Here, Hunyuan-7B-Base$^\star$ denotes the Hunyuan-7B-Base model after MT-oriented Pre-training. The highest and second-best scores are shown in \textbf{bold} and \underline{underlined}, respectively.}}
	\label{tab:pretrain_stage2}
	\begin{tabular}{@{}lcccccc@{}}
		\toprule
		& \multicolumn{2}{c}{\textbf{Flores-200}} & \multicolumn{2}{c}{\textbf{WMT24pp}} & \multicolumn{2}{c}{\textbf{Mandarin$\Leftrightarrow$Minority}}\\
		& XCOMET-XXL & CometKiwi & XCOMET-XXL & CometKiwi & XCOMET-XXL & CometKiwi \\
		\midrule
		Qwen3-8B-Base & \underline{57.88} & \underline{55.46} & \underline{35.89} & \underline{36.69} & \underline{32.02} & \underline{23.98}\\
		Hunyuan-7B-Base$^\star$ & \textbf{67.41} & \textbf{65.87} & \textbf{48.34} & \textbf{46.29} & \textbf{39.95} & \textbf{28.05}\\
		\bottomrule
	\end{tabular}
\end{table}

\section{Discussion}

\subsection{Case Study}
Table~\ref{table: test_scenario_1_4}, Table~\ref{table: test_scenario_5_6}, and Table~\ref{table: test_scenario_Chimera} illustrate representative translation cases, comparing the outputs of Hunyuan-MT-7B and Hunyuan-MT-Chimera-7B on key testing points. These translation cases demonstrate the model's sophisticated contextual understanding and proficiency in managing complex translations.

Case analysis suggests that Hunyuan-MT-7B possesses a notable capacity for handling culturally-specific and context-dependent phenomena that often challenge conventional translation systems. For instance, in the context of Chinese social media, the model correctly interprets "小红薯" as the platform "REDnote" and "砍一刀" as a specific e-commerce promotional feature. In contrast, a baseline system like Google Translate tends to produce literal but contextually incorrect renderings ("sweet potatoes" and "cuts"). This sensitivity to non-literal meaning appears to extend to English idiomatic expressions (e.g., "You are killing me" as amusement) and domain-specific terminology, including medical terms ("blood disorders") and structured data like mailing addresses, which are often left untranslated by the baseline. These instances indicate an ability to leverage world knowledge and pragmatic context beyond direct lexical mapping.

\begin{center}
	\scriptsize
	\renewcommand{\arraystretch}{1.4}
	\setlength{\tabcolsep}{3pt}
	\begin{longtable}{p{3cm}p{12.5cm}}  
		\caption{Comparative case studies across different scenarios. Red text highlights translation errors in Google-Translator outputs.} \\
		\hline
		\multicolumn{2}{c}{\textbf{Scenario 1: Chinese Social}} \\
		\hline
		
		\textbf{Example \#1} & 都知道小红薯在国外疯魔了，数据来告诉你有多逆天！ \\
		\textbf{Testing Points} & “小红薯”: REDnote. \\
		\textbf{Huanyuan-MT-7B} & Everyone knows that \textcolor{blue}{\textbf{REDnote}} has become incredibly popular abroad, and the data is proof of just how remarkable its success is! \\
		\textbf{Google-Translator} & Everyone knows that \textcolor{red}{\textbf{sweet potatoes}} are popular abroad. Here are the statistics to tell you how popular they are! \\
		\hline
		
		\textbf{Example \#2} & 没有一个外国人可以逃脱拼多多砍一刀 \\
		\textbf{Testing Points} & “砍一刀”: Its core mechanism revolves around social sharing to reduce product prices. \\
		\textbf{Huanyuan-MT-7B} & No foreigner can escape the \textcolor{blue}{\textbf{price cuts}} offered by Pinduoduo. \\
		\textbf{Google-Translator} & No foreigner can escape Pinduoduo's \textcolor{red}{\textbf{cuts}} \\
		\hline
		
		\multicolumn{2}{c}{\textbf{Scenario 2: English Slang}} \\
		\hline
		
		\textbf{Example \#1} & Oh!You are killing me! \\
		\textbf{Testing Points} & "You are killing me": it is an informal and exaggerated English expression used to emphasize intense feelings like extreme amusement, frustration, or being overwhelmed. \\
		\textbf{Huanyuan-MT-7B} & 哦！你\textcolor{blue}{\textbf{真要把我笑死了}}！ \\
		\textbf{Google-Translator} & 噢！你\textcolor{red}{\textbf{要杀死我了}}！\\
		\hline
		
		\textbf{Example \#2} & If you need advice on vintage vinyl, talk to Jim; he's the OG collector around here. \\
		\textbf{Testing Points} & "OG": s a slang term originally short for "Original Gangster," now mainly used to describe a senior, authoritative, or pioneering person/thing in a field (with no negative connotation, often a term of recognition).  \\
		\textbf{Huanyuan-MT-7B} & 如果你需要关于复古黑胶唱片方面的建议，可以找吉姆聊聊——他是这里\textcolor{blue}{\textbf{最资深}}的收藏家。 \\
		\textbf{Google-Translator} & 如果您需要有关老式黑胶唱片的建议，请与 Jim 联系；他是这里的\textcolor{red}{\textbf{ OG }}收藏家。 \\
		\hline
		
		\multicolumn{2}{c}{\textbf{Scenario 3: Terminology}} \\
		\hline
		
		\textbf{Example \#1} & 已知有血液疾病及尿酸性肾结石的患者不推荐使用本品，二岁以下儿童不得服用。 \\
		\textbf{Testing Points} & “血液疾病”: blood disorder, “尿酸性肾结石”: uric acid kidney stone. \\
		\textbf{Huanyuan-MT-7B} & This product is not recommended for patients with known \textcolor{blue}{\textbf{blood disorders}} or \textcolor{blue}{\textbf{uric acid kidney stones}}, and it should not be taken by children under the age of two. \\
		\textbf{Google-Translator} & Patients with known \textcolor{red}{\textbf{blood diseases}} and \textcolor{red}{\textbf{uricidal}} kidney stones are not recommended for use, and children under two years of age are not allowed to take it. \\
		\hline
		
		\multicolumn{2}{c}{\textbf{Scenario 4: Place Name}} \\
		\hline
		
		\textbf{Example \#1} & The YMU community mourns the loss of Dr. Wang, a beloved faculty member whose legacy lives on through his students. \\
		\textbf{Testing Points} & "YMU": Yunnan University of Nationalities. \\
		\textbf{Huanyuan-MT-7B} & \textcolor{blue}{\textbf{云南民族大学（YMU）}}的师生们为王博士的离世感到悲痛。他是一位深受爱戴的教师，他的精神将通过他的学生们得以延续。 \\
		\textbf{Google-Translator} & \textcolor{red}{\textbf{杨百翰大学}}全体师生对王博士的去世深感悲痛，王博士是一位受人爱戴的教师，他的精神遗产通过他的学生得以传承。 \\
		\hline
		
		\textbf{Example \#2} & Unit 602, HUANGPU Complex, 654 Huangpu Drive, Huangpu District, Guangzhou, Guangdong Province, 510000, China \\
		\textbf{Huanyuan-MT-7B} & \textcolor{blue}{\textbf{中国广东省广州市黄埔区黄埔大道654号黄埔大厦602单元，邮编510000}} \\
		\textbf{Google-Translator} & \textcolor{red}{\textbf{Unit 602, HUANGPU Complex, 654 Huangpu Drive, Huangpu District, Guangzhou, Guangdong Province, 510000, China}} \\
		\hline
		
		\label{table: test_scenario_1_4}
	\end{longtable}
	
\end{center}

Furthermore, Hunyuan-MT-7B demonstrates promising results across a diverse set of language pairs, including those with fewer digital resources. For translations involving widely spoken European and Asian languages (e.g., Italian, German, Korean, Persian), the model tends to generate more idiomatic and contextually appropriate outputs. Its performance is particularly noteworthy in low-resource language translation. When translating from Mandarin to ethnic minority languages such as Kazakh and Tibetan, Hunyuan-MT-7B is often able to produce coherent sentences where the baseline system may yield nonsensical or failed outputs. This suggests our approach may be particularly beneficial for improving translation quality in low-resource scenarios, a finding that aligns with the quantitative results on our Mandarin⇒Minority benchmark.

\begin{center}
	\scriptsize
	\renewcommand{\arraystretch}{1.6}
	\setlength{\tabcolsep}{3pt}
	\begin{longtable}{p{3cm}p{12.5cm}}  
		\caption{Comparative case studies across different scenarios. Red text highlights translation errors in Google-Translator outputs.} \\
		\hline
		\multicolumn{2}{c}{\textbf{Scenario 5: Less Commonly Used Languages}} \\
		\hline
		
		\textbf{Example \#1} & Questo avviene anche in Norvegia, Svezia e Nuova Zelanda, ma in generale è un fatto piuttosto singolare (ad es. nei Paesi Bassi il rapporto è di uno a quaranta). \\
		\textbf{Testing Points} & Italian $\Rightarrow$ English \\
		\textbf{Huanyuan-MT-7B} & This is also the case in Norway, Sweden, and New Zealand. However, overall it’s a rather \textcolor{blue}{\textbf{unusual phenomenon}}; for example, in the Netherlands the \textcolor{blue}{\textbf{ratio}} is one to forty. \\
		\textbf{Google-Translator} & This also occurs in Norway, Sweden and New Zealand, but in general it is a rather \textcolor{red}{\textbf{singular fact}} (e.g. in the Netherlands the \textcolor{red}{\textbf{relationship}} is one by forty). \\
		\hline
		
		\textbf{Example \#2} & Mehrere Geiseln wurden gerettet und es gibt bisher mindestens sechs bestätigte Tote. \\
		\textbf{Testing Points} & German $\Rightarrow$ English \\
		\textbf{Huanyuan-MT-7B} & Several hostages have been \textcolor{blue}{\textbf{rescued}}, and there are at least six confirmed deaths so far. \\
		\textbf{Google-Translator} & Several hostages have been \textcolor{red}{\textbf{saved}} and there are at least six confirmed deaths so far. \\
		\hline
		
		\textbf{Example \#3} & 
		\adjustbox{valign=t, max width=14.5cm, width=\linewidth, margin=0pt 0pt 0pt 1pt}{\includegraphics{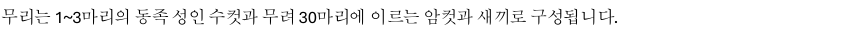}} \\
		\textbf{Testing Points} & Korean $\Rightarrow$ Chinese \\
		\textbf{Huanyuan-MT-7B} & \textcolor{blue}{\textbf{这样的群体}}通常由1到3只同类的成年雄性，以及多达30只雌性和幼崽组成。 \\
		\textbf{Google-Translator} & \textcolor{red}{\textbf{一个狼群}}由一至三只同种成年雄性和最多 30 只雌性及其幼崽组成。 \\
		\hline
		
		\textbf{Example \#4} & 
		\adjustbox{valign=t, max width=14.5cm, width=\linewidth, margin=0pt 0pt 0pt 1pt}{\includegraphics{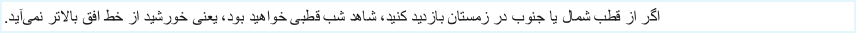}} \\
		\textbf{Testing Points} & Persian $\Rightarrow$ Chinese \\
		\textbf{Huanyuan-MT-7B} & 如果您在冬季前往北极或南极，将会目睹\textcolor{blue}{\textbf{极夜现象}}，也就是说太阳始终无法升出地平线。 \\
		\textbf{Google-Translator} & 如果您在冬季参观北极或南极，您会看到一个\textcolor{red}{\textbf{极地夜晚}}，这意味着太阳不会高于地平线。 \\
		\hline
		
		\multicolumn{2}{c}{\textbf{Scenario 6: Ethnic Minority Languages}} \\
		\hline
		
		\textbf{Example \#1} & 
		\adjustbox{valign=t, max width=14.5cm, width=\linewidth, margin=0pt 0pt 0pt 1pt}{\includegraphics{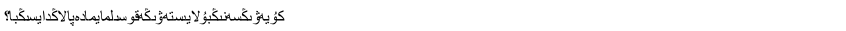}} \\
		\textbf{Testing Points} & Kazakh $\Rightarrow$ Chinese \\
		\textbf{Huanyuan-MT-7B} & \textcolor{blue}{\textbf{你担心你的丈夫不同意你这样做吗？}} \\
		\textbf{Google-Translator} & \textcolor{red}{\textbf{不欢迎Kieuingulets吗？}} \\
		\hline
		
		\textbf{Example \#2} & 
		\adjustbox{valign=t, max width=14.5cm, width=\linewidth, margin=0pt 0pt 0pt 1pt}{\includegraphics{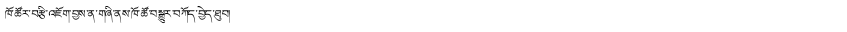}} \\
		\textbf{Testing Points} & Tibetan $\Rightarrow$ Chinese \\
		\textbf{Huanyuan-MT-7B} & 只有尊重他们，才能\textcolor{blue}{\textbf{真正改变他们。}} \\
		\textbf{Google-Translator} & 只有尊重它们，才能\textcolor{red}{\textbf{进行改革。}} \\
		\hline
		
		\textbf{Example \#3} & 
		黄尸癫狗又出黎周围吠人了! \\
		\textbf{Testing Points} & Cantonese => Chinese \\
		\textbf{Huanyuan-MT-7B} & 黄尸\textcolor{blue}{\textbf{疯狗又出来到处}}吠人了！ \\
		\textbf{Google-Translator} & 黄尸\textcolor{red}{\textbf{癫狗又出黎周围}}吠人了! \\ 
		\hline
		
		\textbf{Example \#4} & 
		\adjustbox{valign=b, max width=14.5cm, width=\linewidth, margin=0pt 1pt 0pt 0pt}{\includegraphics{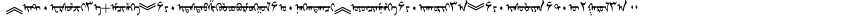}} \\
		\textbf{Testing Points} & Mongolian => Chinese \\
		\textbf{Huanyuan-MT-7B} & \textcolor{blue}{\textbf{提升“互联网+医疗”水平，内蒙古首家“智慧医院”上线。}} \\
		\textbf{Google-Translator} & 
		\adjustbox{valign=b, max width=14.5cm, width=\linewidth, margin=0pt 1pt 0pt 1pt}{\includegraphics{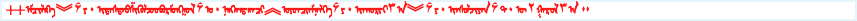}}  \\
		\hline
		
		\textbf{Example \#5} & 
		\adjustbox{valign=b, max width=14.5cm, width=\linewidth, margin=0pt 1pt 0pt 0pt}{\includegraphics{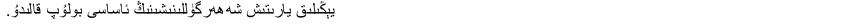}} \\
		\textbf{Testing Points} & Uyghur => Chinese \\
		\textbf{Huanyuan-MT-7B} & 创新将成为\textcolor{blue}{\textbf{城市繁荣发展}}的基础。 \\
		\textbf{Google-Translator} & 创新是\textcolor{red}{\textbf{城镇化}}的基础。 \\
		\hline

		\label{table: test_scenario_5_6}
	\end{longtable}
\end{center}

Building upon this foundation, the Hunyuan-MT-Chimera-7B variant exhibits further refinements in nuanced interpretation. This is exemplified by its ability to resolve ambiguities in specialized domains, such as identifying "d2" as the game "Diablo II" and interpreting "make a game" as a term for in-game trading. The model also shows a sophisticated approach to informal language, pragmatically translating the intensifier "fucking" to convey emphasis rather than resorting to a literal profanity. In other scenarios, it correctly maps "三分" to "three points" in a sports context. It demonstrates greater situational awareness by rendering "穿过" as "sped through" a crowd, avoiding the problematic implication of "drove through." These cases suggest that the Chimera enhancement facilitates a more fine-grained understanding of colloquialisms and domain-specific context, contributing to translations that are both more accurate and pragmatically sound.

\begin{center}
	\scriptsize
	\renewcommand{\arraystretch}{1.3} 
	\setlength{\tabcolsep}{3pt} 
	\begin{longtable}{p{2cm}p{13.5cm}} 
		\caption{Case studies of the Hunyuan-MT-Chimera-7B model.} \label{table: test_scenario_Chimera} \\
		\hline
		\multicolumn{2}{c}{\textbf{Scenario 7: Hunyuan-MT-Chimera-7B Case Study}} \\
		\hline
		\textbf{Example \#1} & I recently returned to d2 after several year, now I’m windering:\newline Where do you guys sell/buy your stuff? Do you just make a game “O xxx N yyy” and hope for the best? Or are there a website that’s more efficient? \\
		\textbf{Testing Points} & "d2": Diablo II, "make a game “O xxx N yyy”": it means creating a game room where the room name uses "O" (short for "Offer," standing for the item xxx you want to sell) and "N" (short for "Need," standing for the item yyy you want in exchange) to inform other players of your trading needs. \\
		\textbf{Model \#1} & 我最近几年后又回到了\textcolor{red}{\textbf{d2}}，现在我在想：\newline 你们一般在哪里买卖装备？你们只是开个游戏房间“O xxx N yyy”，然后碰运气吗？还是有更高效的网站？ \\
		\textbf{Model \#2} & 我最近在几年后重新回到了\textcolor{red}{\textbf{d2}}，现在我想知道：\newline 你们都在哪里买卖东西？你们是直接创建一个"O xxx N yyy"的游戏然后碰运气吗？还是有更高效的网站？ \\
		\textbf{Model \#3} & 我最近时隔几年又回来玩\textcolor{red}{\textbf{d2}}了，想问一下：\newline 各位都是在哪里交易物品的？是自己建个名叫“\textcolor{red}{\textbf{出xxx换yyy}}”的房间然后随缘等，还是有更高效的网站？ \\
		\textbf{Model \#4} & 时隔多年重回暗黑2，现在我想知道：\newline 你们都在哪买卖装备？是直接建个“\textcolor{red}{\textbf{O物品换N符文}}”的房间碰运气吗？还是有更高效的交易网站？ \\
		\textbf{Model \#5} & 我最近在离开几年后重新回到\textcolor{red}{\textbf{d2}}，现在我很疑惑：\newline 你们都是在哪里买卖东西的？你们只是\textcolor{red}{\textbf{做一款游戏}}叫“O xxx N yyy”，然后听天由命吗？还是有更高效的网站？ \\
		\textbf{Model \#6} & 我最近在几年后回到了\textcolor{red}{\textbf{d2}}，现在我在想：\newline 你们\textcolor{red}{\textbf{这些人}}在哪里卖/买你们的东西？你们只是\textcolor{red}{\textbf{做一个游戏}}“O xxx N yyy”然后\textcolor{red}{\textbf{希望是最好的}}吗？还是有一个更有效率的网站？ \\
		\textbf{Hunyuan-MT-\newline Chimera-7B} & 时隔多年，我又重新开始玩\textcolor{blue}{\textbf{《暗黑破坏神2》（Diablo II）}}，现在我想知道：\newline 大家都是在哪里进行物品交易的？是通过自己\textcolor{blue}{\textbf{创建名为“O xxx N yyy”的游戏房间来交易}}，然后\textcolor{blue}{\textbf{听天由命}}吗？还是有更高效的交易网站？ \\
		\hline
		
		\textbf{Example \#2} & Now I obviously can’t get out and shut it on the main road, so I pull down a few side streets and ensure that no one fucking knows I’ve quickly jumped out and shut it.I jump back in and drive home like nothing’s happened. \\
		\textbf{Testing Points} & "no one fucking knows": it means that absolutely no one is aware of, with the word "fucking" used here as a strong informal intensifier to emphasize the degree of secrecy—stressing that the action is completely unknown to anyone. \\
		\textbf{Model \#1} & 显然我不能在主路上下车去关门，所以我驶进几条小路，确保没人\textcolor{red}{\textbf{他妈的}}知道我迅速下车把门关上了。我跳回车里，像什么都没发生一样开车回家。 \\
		\textbf{Model \#2} & 我显然不能在主路上下车去关门，所以我拐进了几条小巷，确保没有人知道我快速跳下车关了门。我跳回车里，像什么都没发生一样开车回家。 \\
		\textbf{Model \#3} & 我显然不能在主路上停车关门，于是我拐进几条小路，确保\textcolor{red}{\textbf{他妈的}}没人知道我赶紧跳下车把门关上。我再跳回车里，若无其事地开车回家。 \\
		\textbf{Model \#4} & 现在我显然没法在大马路上停车关门，所以我拐进几条小巷，确保没人\textcolor{red}{\textbf{他妈}}发现我迅速跳下车关好门。我跳回车里，若无其事地开回家。 \\
		\textbf{Model \#5} & 现在我显然不能在主干道上停下来关门，所以我拐进几条小街，确保没人\textcolor{red}{\textbf{他妈的}}知道我迅速跳下车把门关上。我跳回车上，像什么都没发生一样开车回家。 \\
		\textbf{Model \#6} & 显然我不能在主路上下车去关门，所以我驶进几条小路，确保没人\textcolor{red}{\textbf{他妈的}}知道我迅速下车把门关上了。我跳回车里，像什么都没发生一样开车回家。  \\
		\textbf{Hunyuan-MT-\newline Chimera-7B} & 显然我不能在主路上停车、下车去关门；于是我把车拐进几条小巷，确保\textcolor{blue}{\textbf{没有人看到}}我迅速下车把门关上。之后我再跳回车里，若无其事地开车回家。 \\
		\hline
		
		\textbf{Example \#3} & 希望我们能得到三分，不过那将是一场棘手的比赛，因为他们在上一场比赛中取得了不错的成绩，但是有这么多人支持我们，我相信我们一定可以继续前进，取得好成绩。 \\
		\textbf{Testing Points} & "三分": three points, not for winning three of the matches \\
		\textbf{Model \#1} & Hopefully, we can \textcolor{red}{\textbf{win all three of those matches}}. It’s going to be a tough competition, though, because they performed well in their last match. But with so many people supporting us, I believe we can continue to do well and achieve good results. \\
		\textbf{Model \#2} & Hopefully, we can manage to \textcolor{red}{\textbf{win three of these matches}}. It’s going to be a tough competition, though, because they performed well in their last match. But with so many people supporting us, I believe we can keep going and achieve good results. \\
		\textbf{Model \#3} & Hopefully, we can manage to \textcolor{red}{\textbf{win three of these matches}}. It’s going to be a tough challenge, though, because they performed well in their last match. But with so many people supporting us, I’m confident that we can continue to move forward and \textcolor{red}{\textbf{do well}}. \\
		\textbf{Model \#4} & Hopefully, we can secure three points. It’s going to be a tough match, because they performed well in their last game. However, with so many people supporting us, I believe we can continue to move forward and \textcolor{red}{\textbf{do well}}. \\
		\textbf{Model \#5} & Hopefully, we can secure three points. It’s going to be a tough match, though; they performed well in their last game. But with so many people supporting us, I’m confident that we can continue to move forward and do well. \\
		\textbf{Model \#6} & Hopefully, we can secure three points. It’s going to be a tough match, though; they performed well in their last game. But with so many people supporting us, I believe we can continue to move forward and \textcolor{red}{\textbf{do well}}.  \\
		\textbf{Hunyuan-MT-\newline Chimera-7B} & I hope we can secure those \textcolor{blue}{\textbf{three points}}, but it’s going to be a tough match, as they performed well in their last game. However, with so many people supporting us, I’m confident that we can continue to move forward and \textcolor{blue}{\textbf{achieve good results}}. \\
		\hline
		
		\textbf{Example \#4} & 一辆 1948 年的福特水星汽车穿过佐治亚州门罗小镇的一群围观者，朝着小小的摩尔滩桥隆隆奔行。 \\
		\textbf{Model \#1} & A 1948 Ford Mercury \textcolor{red}{\textbf{drove through}} a crowd of onlookers in the town of Monroe, Georgia, and then made its way \textcolor{red}{\textbf{across}} the small Moorestown Bridge. \\
		\textbf{Model \#2} & A 1948 Ford Mercury \textcolor{red}{\textbf{drove through}} a crowd of onlookers in the town of Monroe, Georgia, and then made its way \textcolor{red}{\textbf{across}} the small Moore’s Bay Bridge. \\
		\textbf{Model \#3} & A 1948 Ford Mercury \textcolor{red}{\textbf{convertible drove through}} a crowd of onlookers in Monroe, Georgia, and then made its way \textcolor{red}{\textbf{across}} the small Moore’s Bend Bridge. \\
		\textbf{Model \#4} & A 1948 Ford Mercury \textcolor{red}{\textbf{drove through}} a crowd of onlookers in the town of Monroe, Georgia, and then made its way \textcolor{red}{\textbf{across}} the small Moorestown Bridge. \\
		\textbf{Model \#5} & A 1948 Ford Mercury \textcolor{red}{\textbf{driven by a man named George White}} sped through a crowd in Monroe, Georgia, and then over the small Moore’s Bay Bridge. \\
		\textbf{Model \#6} & A 1948 Ford Mercury \textcolor{red}{\textbf{driven by a man named Steve Myers}} sped through a crowd in Monroe, Georgia, and over the small bridge that crosses Moore Creek.  \\
		\textbf{Hunyuan-MT-\newline Chimera-7B} & A 1948 Ford Mercury \textcolor{blue}{\textbf{sped through}} a crowd of onlookers in the small town of Monroe, Georgia, \textcolor{blue}{\textbf{on its way towards}} the tiny Moorestown Bridge. \\
		\hline
	\end{longtable}
\end{center}

\subsection{The Role of CoT in MT}
We also investigate the integration of Chain-of-Thought (CoT) methodologies within our training framework for translation. Our experiments indicate that when the reward signal is applied exclusively to the final translation output, it proves insufficient to elicit a meaningful reasoning process. Under this condition, the model tends to generate generic, boilerplate statements for the CoT, such as, "I need to translate the English text into Chinese and ensure the translation accurately conveys the original meaning." Consequently, this approach yields no discernible improvement in translation performance when compared to a baseline model trained without CoT.

In contrast, we explore an alternative strategy that provides reward signals for both the reasoning process itself and the final translation, a method consistent with our work on TAT-R1 \citep{TAT-R1}. We find that this dual-reward structure successfully incentivizes the model to produce a more substantive and task-relevant CoT. This, in turn, correlates with measurable improvements in overall translation quality.

\subsection{Human Evaluation}
\textbf{Hunyuan-MT Challenge Testset}, which encompasses a diverse range of scenarios, including social interactions, emails, food ordering, shopping, and navigation inquiries. This test set involves bidirectional translation between Chinese and English. Existing open-source test sets predominantly focus on the news domain. To comprehensively evaluate the translation capabilities, we construct a challenging test set that covers multiple domains, including news, medicine, government, literature, law, natural sciences, arts, computing, and the internet.

As presented in Table~\ref{tab:human_evaluation}, the experimental results reveal distinct performance tiers among the six evaluated translation models. The top-performing cluster, comprising Gemini-2.5-Pro (3.223), DeepSeek-V3-0324 (3.219), and Huanyuan-MT-7B (3.189), demonstrates marginal performance differences of less than 0.034 points, suggesting a convergence of state-of-the-art translation capabilities. Notably, Gemini-2.5-Pro exhibits exceptional bidirectional balance with nearly identical scores for ZH$\Rightarrow$EN (3.225) and EN$\Rightarrow$ZH (3.222) translations. 

\begin{table}[ht!]
	\centering
	\renewcommand{\arraystretch}{1.1}
	\setlength{\tabcolsep}{11pt}
	\caption{Human evaluation of translation quality for the Chinese-to-English (ZH$\Rightarrow$EN) and English-to-Chinese (EN$\Rightarrow$ZH) directions.}
	\label{tab:human_evaluation}
	\begin{tabular}{lccc}
		\toprule
		\textbf{Models} & \textbf{ZH$\Rightarrow$EN} & \textbf{EN$\Rightarrow$ZH} & \textbf{Avg.} \\
		\midrule
		Gemini-2.5-Pro & 3.225 & 3.222 & 3.223 \\
		DeepSeek-V3-0324 & 3.253 & 3.203 & 3.219 \\
		Qwen3-32B & 3.137 & 3.073 & 3.094 \\
		Google-Translator & 2.841 & 2.101 & 2.344 \\
		Seed-X-PPO-7B & 3.139 & 3.033 & 3.068 \\
		Huanyuan-MT-7B & 3.258 & 3.155 & 3.189 \\
		\bottomrule
	\end{tabular}
\end{table}

At the same time, most other models display a systematic bias favoring Chinese-to-English translation over the reverse direction, a phenomenon potentially attributable to the greater complexity of generating grammatically correct Chinese text. The significant performance gap between these leading models and Google-Translator (2.344), which underperforms by approximately 27\%, underscores the superiority of modern transformer-based architectures over traditional translation systems. Particularly noteworthy is the competitive performance of Huanyuan-MT-7B, which, despite its relatively modest 7B parameters, achieves results comparable to larger models, suggesting that task-specific optimization can effectively compensate for model scale in specialized translation tasks.

\section{Conclusion}
In this report, we introduce Hunyuan-MT-7B and Hunyuan-MT-Chimera-7B, a family of open-source LLMs specifically engineered for machine translation, enabling bidirectional translation across 33 languages. We present a comprehensive overview of our training methodology, which encompasses pre-training, supervised fine-tuning, and reinforcement learning phases. Throughout this exposition, we share critical insights and best practices gleaned from our iterative optimization process.
Remarkably, with only 7B parameters, Hunyuan-MT-7B and Hunyuan-MT-Chimera-7B achieve translation quality that rivals—and in certain cases surpasses—that of state-of-the-art LLMs and leading commercial translation systems, as validated through both automatic metrics and human evaluation. By making Hunyuan-MT's model weights publicly available, we aim to empower the research community with an accessible, high-performance foundation model that can accelerate innovation in MT research and applications.

\section{Contributions and Acknowledgments}
\subsection{Core Contributors}
Mao Zheng, Zheng Li, Bingxin Qu, Mingyang Song, Yang Du, Mingrui Sun, Di Wang

\subsection{Contributors}
Tao Chen, Jiaqi Zhu, Xingwu Sun, Yufei Wang, Can Xu, Chen Li, Kai Wang, Decheng Wu

\subsection{Acknowledgments}
We thank Shen Huang for providing the training data and sharing valuable insights on Mandarin$\Leftrightarrow$Minority translation.

\end{CJK*}
\bibliography{colm2024_conference}
\bibliographystyle{colm2024_conference}
\end{document}